\documentclass{article}

\usepackage{arxiv}

\usepackage[utf8]{inputenc} 
\usepackage[T1]{fontenc}    
\usepackage{hyperref}       
\usepackage{url}            
\usepackage{booktabs}       
\usepackage{amsfonts}       
\usepackage{nicefrac}       
\usepackage{microtype}      
\usepackage{lipsum}
\usepackage{graphicx}
\usepackage{amssymb}
\usepackage{amsmath}

\usepackage[english]{babel}
\usepackage[dvipsnames, svgnames, x11names]{xcolor}
\usepackage{graphicx}
\usepackage{amsmath}
\usepackage{amssymb}
\usepackage{booktabs}
\usepackage{multirow}
\usepackage{xcolor}         
\usepackage{colortbl}
\usepackage{xspace}
\usepackage[ruled,vlined]{algorithm2e}
\usepackage{wrapfig}
\usepackage{booktabs}
\usepackage{rotating}

\title{S2A: A Unified Framework for Parameter and Memory Efficient Transfer Learning}

\author{
 Tian Jin ,Enjun DU ,Changwei Wang ,Wenhao Xu ,Ding Luo \\
  Sichuan University,  TongJi University\\
  Shandong Computer Science Center, Beijing University of Posts and Telecommunications\\
   The Hong Kong University of Science and Technology (Guangzhou)\\
  \texttt{2022141520260@stu.scu.edu.cn}    \texttt{2452687@tongji.edu.cn} \\
   \And
}

\begin{document}
\maketitle
\begin{abstract}

Parameter-efficient transfer learning (PETL) aims to reduce the scales of pretrained models for multiple downstream tasks. However, as the models keep scaling up, the memory footprint of existing PETL methods is not significantly reduced compared to the reduction of learnable parameters. This limitation hinders the practical deployment of PETL methods on memory-constrained devices. To this end, we proposed a new PETL framework, called Structure to Activation (S2A), to reduce the memory footprint of activation during fine-tuning. Specifically, our framework consists of: 1) Activation modules design(i.e., bias, prompt and side modules) in the parametric model structure, which results in a significant reduction of adjustable parameters and activation memory; 2) 4-bit quantization of activations based on their derivatives for non-parametric structures (e.g., nonlinear functions), which maintains accuracy while significantly reducing memory usage. Our S2A method consequently offers a lightweight solution in terms of both parameters and memory footprint. We evaluated S2A with different backbones and performed extensive experiments on various datasets to evaluate the effectiveness. The results show that our methods not only outperform existing PETL techniques, achieving a fourfold reduction in GPU memory footprint on average, but also shows competitive performance in accuracy with fewer tunable parameters. These demonstrate that our method is highly suitable for practical transfer learning on hardware-constrained devices.

\end{abstract}

\section{Introduction}
Transfer learning has become a prevalent paradigm in computer vision research, where pre-trained models are fine-tuned on target datasets to adapt to specific tasks.~\cite{zhang2025parameter,Chen_2024_CVPR,Zhou_2024_CVPR,zhang2025handling,DSVC}
With the increasing scale of pre-trained models, the number of parameters has grown significantly, making the full fine-tuning less affordable, especially for recent very large models, \emph{e.g.}, ViT-G~\cite{zhai2022scaling} and Swin-G~\cite{liu2022swin}.
Moreover, a single pre-trained model may be transferred to multiple tasks, and each fine-tuned model requires considerable space to store. To reduce the offline storage cost, \emph{Parameter Efficient Transfer Learning} (PETL) is proposed to perform partial fine-tuning on either the original or some specialized additional parametric modules. The additional modules are designed to be lightweight and inserted into the original model during fine-tuning, so that we can update the parameters of such small modules while fixing the original parameters. The representative PETL methods include bias tuning~\cite{zaken2021bitfit} (fine-tuning the bias vectors only), LORA~\cite{hu2021lora} (decomposing weight into two low-rank matrices) and Adapter~\cite{houlsby2019parameter} (inserting parameter-efficient modules into the original model) and VPT~\cite{jia2022visual} (appending learnable tokens into ViTs).In practice, the PETL methods are particularly suitable for cloud-based service where we may require a single model to adapt to multiple streaming tasks without changing pre-trained parameters. 

\begin{figure}[t]
\centering
\includegraphics[width=1\textwidth]{    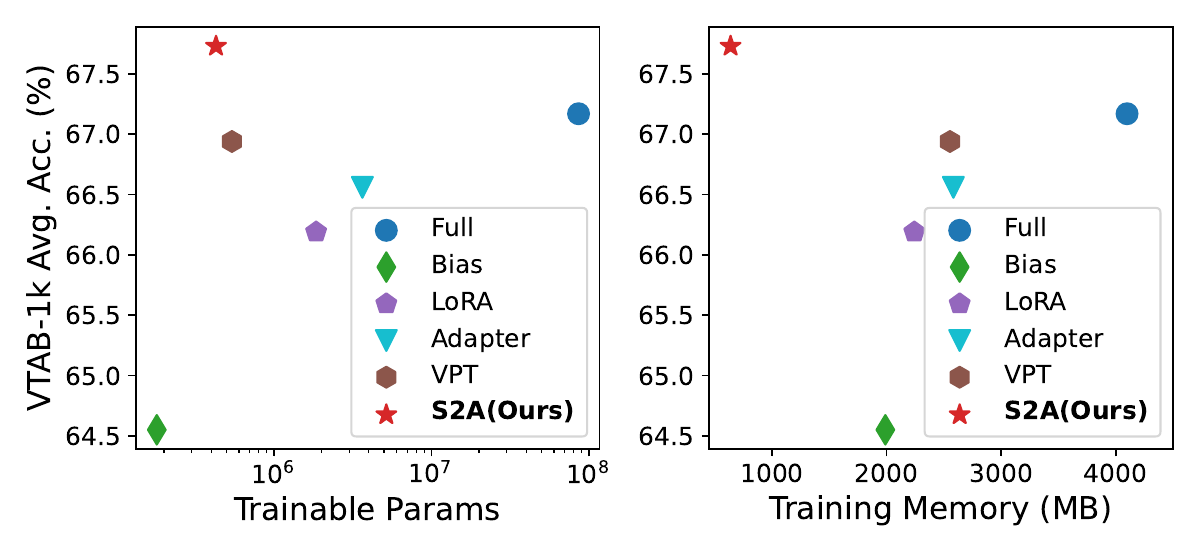}
\caption{VTAB-1k average accuracy v.s. trainable parameters (\textbf{left}) and memory footprint (\textbf{right}) of ViT-B/16. Our method shows appealing performance with fewer learnable parameters and lower memory footprint.}
\label{fig:mem_param}
\end{figure}

However, We note a major limitation of the popular PETL methods - the memory inefficiency. Though the offline storage cost is reduced proportionally to the reduction of learnable parameters, 
We observe that the GPU memory footprint is not decreased significantly as that of learnable parameters in the training stage. For example, as shown in Figure~\ref{fig:mem_param}, the parameters of ViT-B/16 are notably reduced by 99\% while the training-time GPU memory footprint is only reduced by 40\%, compared to the full fine-tuning. As the updated parameters are contained within the backbone language models, in order to calculate gradients for these parameters for backpropagation, it's necessary to run the backward pass through the large pre-trained language models. This prevents PETL methods to many of real-world applications with limited computational resources.

Therefore, we propose a memory-efficient PETL framework that facilitates both training and practical deployment. 
In training stage, the largest memory footprint is shown from the activation (a.k.a feature embeddings) utilized for gradient calculations~\cite{chen2016training,yang2022da3,cai2020,sohoni2019low}.
We reduce the memory of activation by analyzing the training process of both parametric modules and non-parametric modules. For parametric modules, we propose activation-efficient modules which use less activation to update learnable parameters. For non-parametric modules, we propose quantization for the required activation based on their derivative. Our contributions can be summarized as follows:
\begin{itemize}
    \item We introduce S2A, a PETL framework to reduce the training memory of activation by analyzing the forward and backward pass of both parametric modules and non-parametric modules.
    \item We propose Low-Rank Prompt (LRP) and Lite Side Branch (LSB) as activation-efficient parametric modules for transfer learning.
    \item We propose activation quantization for non-parametric modules based on their derivative. It reduces the memory of activation with negligible influence on accuracy (e.g., $\pm 0.4$\% on VTAB-1k and FGVC datastes).
\end{itemize}

\section{Related Works}

\subsection{Parameter-efficient Transfer Learning}
Transfer learning targets adapting the pre-trained model to downstream tasks, for which the most straightforward way is fully fine-tuning all the parameters. However, there are two drawbacks to full fine-tuning. First, it may suffer from catastrophic forgetting \cite{french1999catastrophic,mccloskey1989catastrophic} and over-fitting on relatively small datasets. Second, it also requires lots of storage to save the full copy of the fine-tuned parameters on each downstream dataset. PETL methods are proposed to address those problems. They usually insert extra learnable modules and keep the rest of the model frozen in fine-tuning. In the natural language process(NLP), following the idea of Rebuffi et al.~\cite{rebuffi2018efficient,rebuffi2017learning}, Houlsby et al.~\cite{houlsby2019parameter} propose adapters that are inserted into the middle layers of the model. Each adapter consists of a downsample layer, a non-linear layer, and a upsample layer to modify the hidden states of the pre-trained model. Hu et al.~\cite{hu2021lora} propose LoRA which decompose the linear layer in self-attention into the pre-trained weight and two learnable low-rank weights. After fine-tuning, LoRA merges those low-rank weights to the backbone, thus resulting in no extra computational burden. Meanwhile, Qian el al.~\cite{QIAN2024112058} develop feature separation for indirect diagnosis transfer. In addition to adding extra layers, prompt tuning~\cite{liu2021pre,li2021prefix,lester2021power,liu2021p} add learnable tokens to the input space and don't change the model's architecture. Following the progress in NLP, some PETL methods targeting computer vision(CV) are introduced. With the above methods, the storage for saving tunable parameters is greatly reduced. However, we found the reduction of training memory is not significant as the tunable parameters, limiting the practice on the memory-constraint device. 

Moreover, some recent works emphasize the requirements of both parameter and memory efficiency in the training process\cite{gui2024g,ansell2024scaling}. Ladder Side Tuning (LST)~\cite{sung2022lst} constructs a lightweight side network by keeping the same structure as the pre-trained network but reducing the dimension of each original layer by a predefined reduction factor. In addition, Universal Parallel Tuning (UniPT)~\cite{Diao2024UPT} facilitates the transfer process via a lightweight and learnable parallel network. Nevertheless, we argue that these designs have several potential drawbacks: The LST complexity of the side network goes linearly proportional to the original pre-trained network, making its efficiency susceptible to the original architecture. The UniPT still has a performance gap with fully fine-tuning, and large input sizes may affect the computational complexity. As we mentioned, our S2A outperforms their solutions in terms of adaptability and accuracy, displaying more powerful capabilities and broad applicability over various model architectures in multiple tasks.

\subsection{Neural Network Quantization}
Neural Network Quantization(NNQ)\cite{jacob2018quantization,han2015deep,huang2024towards,choi2018pact,banner2019post,lin2016fixed,novikov2023few,wang2025neural,wei2024advances,li2025scedit,du2025graphmaster,du2025graphoracle} has been widely explored as a method of compression and acceleration. In the training-aware quantization, NNQ quantizes the full-precision (32-bit) weight and activation to lower bit-width before forward calculating. In the backward, the gradient of full-precision weight and activation is substituted by the gradient of the quantized counterpart\cite{yin2019understanding,li2024contemporary}. To discretize the floating-point data, the most direct method is asymmetrical quantization\cite{jacob2018quantization}. In this method, the 32-bit data is projected to the discrete set $\{0,1,2,...,2^N-1\}$ by the scale and shift factor where $N$ is the bit-width. In addition to quantizing the floating data with uniform intervals, there are many works toward non-uniform quantization. Han et al.~\cite{han2015deep} quantizes the weights for less storage by clustering the floating data into different clusters. APoT\cite{li2019additive} uses additive powers-of-two to represent the floating data. Zhao et al.~ \cite{zhao2021distribution} quantizes the weights to non-uniform forms by considering prior distribution. Unlike the previous works that utilize quantization to reduce the storage of weights or speed up the computation in inference stage, our primary goal is to reduce the memory footprint of activation in fine-tuning stage. In our work, we quantize the activation of non-parametric layers based on the properties of their derivative, which has not been explored in NNQ.

\begin{figure*}[t]
\centering
\includegraphics[width=1\linewidth]{    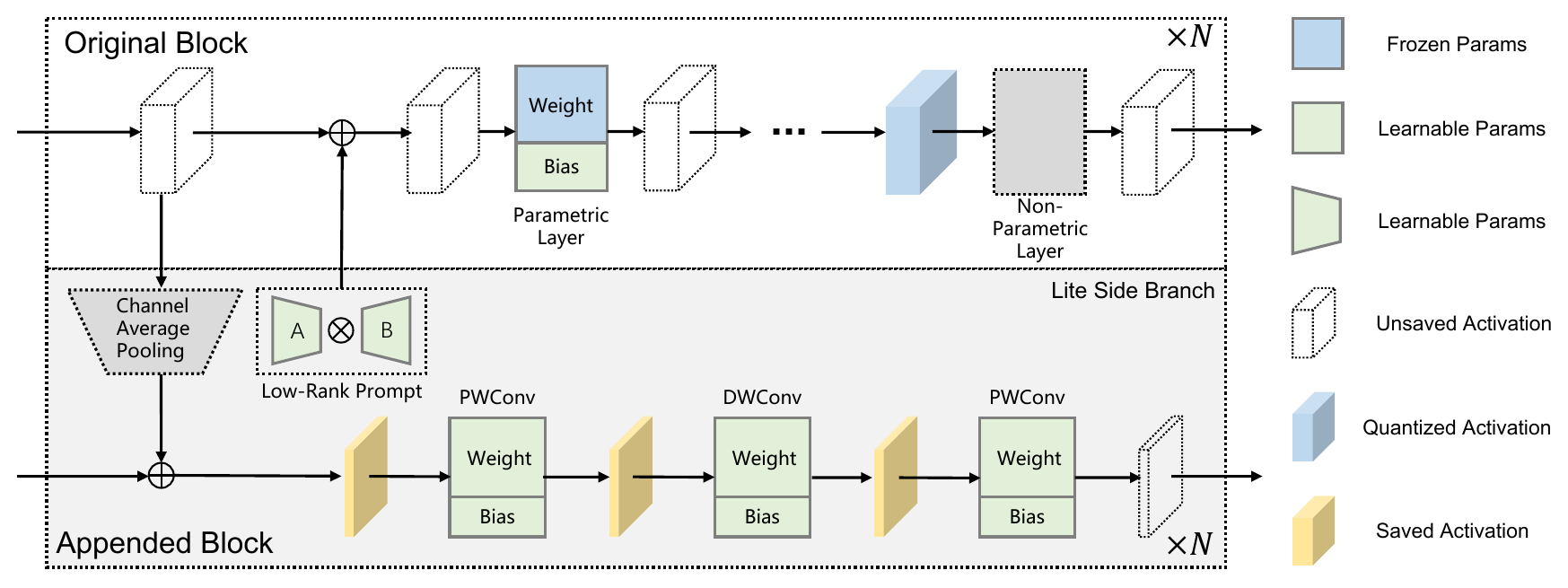}
\caption{Overview of our S2A framework. An original block of the pre-trained model is shown above, and our appended structure is shown below. In the original block, we only tune the bias thus the activation of parametric layers can be released. We quantize the necessary activation of non-parametric layers by analyzing their derivatives. In our appended block, we design activation-efficient Low-Rank Prompt( LRP) and Lite Side Branch (LSB). The last outputs of the two parallel blocks are concatenated for classification.}
\label{fig:pipeline}
\end{figure*}
\vspace{-5pt}

\section{Proposed Method}
In this work, we further develop PETL by reducing the memory of activation from two perspectives: First, We carefully design activation-efficient parametric modules. Second, we quantize the activation of non-parametric layers in the pre-trained model based on their derivatives. Figure~\ref{fig:pipeline} shows an overview of our proposed S2A framework.

\subsection{Parametric Modules}
\label{sec:efficient_modules}
A neural network is composed of a sequence of layers that can be divided into parametric layers and non-parametric layers depending on whether there are parameters in the function or not. We reduce the memory footprint from the activations by considering backpropagation. 

\vspace{-5pt}

\paragraph{Linear layers}
A linear layer with weight $\mathbf{W}$ and bias $\mathbf{b}$ can be formulated as follows:
\begin{equation}
    \mathbf{y} = \mathbf{W} \mathbf{x} + \mathbf{b},
    \label{func:linear_forward}
\end{equation}
where $\mathbf{x}$, $\mathbf{y}$ is the input and output. In the backward pass, we need to calculate the gradient of $\mathbf{x}$, $\mathbf{W}$ and $\mathbf{b}$ for backpropagation and parameter update, which can be written as:
\begin{equation}
    \frac{\partial \mathcal{L}}{\partial \mathbf{x}}= \mathbf{W}^{T} \frac{\partial \mathcal{L}}{\partial \mathbf{y}},\quad
    \frac{\partial \mathcal{L}}{\partial \mathbf{W}} = \frac{\partial \mathcal{L}}{\partial \mathbf{y}}\mathbf{x}^{T}, \quad \frac{\partial \mathcal{L}}{\partial \mathbf{b}} = \frac{\partial \mathcal{L}}{\partial \mathbf{y}}.
    \label{func:linear_backward}
\end{equation}
According to equation~\eqref{func:linear_backward}, the activation of $\mathbf{x}$ is only necessary when $\mathbf{W}$ needs an update. Thus, we freeze $\mathbf{W}$ and only tune the $\mathbf{b}$ of the pre-trained model during fine-tuning to exclude the activation of $\mathbf{x}$. We note that this design also suits convolution and normalization layers since they can be regarded as special types of linear layers. However, only fine-tuning the bias is not enough to retain the performance (see bias tuning in Figure~\ref{fig:mem_param}). We propose learnable modules to prompt the embedding in a memory-efficient way.

\vspace{-5pt}

\paragraph{Low-Rank Prompt (LRP)}
We propose LRP to modify the inputs of intermediate layers by two learnable low-rank matrices. For the activation $\mathbf{x}_i \in \mathbb{R}^{N \times C}$, we add an extra  matrix with the same shape which can be decomposed into two learnable light-weight matrices $\mathbf{A} \in \mathbb{R}^{N \times r}$ and $\mathbf{B} \in \mathbb{R}^{r \times C}$ where $r << N$ and $r << C$ such that
\begin{equation}
    \mathbf{y} = \mathbf{x} + \mathbf{A} \mathbf{B}.
    \label{func:lrp_forward}
\end{equation}
We initialize $\mathbf{A}$ to 0 and $\mathbf{B}$ to $\mathcal{N}(0,\sigma ^2)$ to make the activation fixed at the beginning of fine-tuning. The backward pass of LRP is formulated as,
\begin{equation}
    \frac{\partial \mathcal{L}}{\partial \mathbf{x}} = \frac{\partial \mathcal{L}}{\partial \mathbf{y}}, \quad \frac{\partial \mathcal{L}}{\partial \mathbf{A}} = \frac{\partial \mathcal{L}}{\partial \mathbf{y}} \mathbf{B}^T,\quad \frac{\partial L}{\partial \mathbf{B}} = \mathbf{A}^T \frac{\partial \mathcal{L}}{\partial \mathbf{y}}.
    \label{func:lrp_backward}
\end{equation}
equation~\eqref{func:lrp_backward} shows that the gradients are calculated by $\frac{\partial \mathcal{L}}{\partial \mathbf{y}}$ , $\mathbf{A}$ and $\mathbf{B}$. These terms are independent of the activations $\mathbf{x}$ and $\mathbf{y}$. So we can release the memory of $\mathbf{x}$ once $\mathbf{y}$ is obtained for memory-efficient design.

\paragraph{Lite Side Branch (LSB)}
We propose LSB to utilize the hidden features by adding extra lightweight blocks parallel to the main branch. Specifically, the features of the main branch are first downsampled  and then fed into a lightweight block which is composed of two point-wise and one depth-wise convolution. To downsample the features from the main branch, the common choice is to use a parametric linear layer such as fully-connected or convolution layers. We propose Channel Average Pooling(CAP) as a connector between the main branch and the lite side branch for parameter and memory efficiency. For an input $\mathbf{x} \in \mathbb{R}^{N\times C}$ which will be downsampled to $\mathbf{y}\in \mathbb{R}^{N \times  \frac{C}{r}}$ where $r$ is a compression factor. The forward and backward of CAP can be written as, 
\begin{equation}
    \mathbf{y}_k = \frac{1}{r} \sum_{j=0}^{r-1} \mathbf{x}_{k\times r + j}, \quad     \frac{\partial \mathcal{L}}{\partial \mathbf{x}_{k\times r + j}} = \frac{\partial \mathcal{L}}{\partial \mathbf{y}_k} \frac{1}{r}
    \label{func:cap_fullpass}
\end{equation}
where $k$ is the index in $\{1,2,...,\frac{C}{r}\}$. equation~\eqref{func:cap_fullpass} shows the CAP is parameter-free and the gradient of input $\mathbf{x}$ is only related to the gradient of output $\mathbf{y}$ (i.e., $\frac{\partial \mathcal{L}}{\partial \mathbf{y}}$) and $r$, thus the activation $\mathbf{x}$ can be released after computing the forward part of equation~\eqref{func:cap_fullpass}. The downsampled features $\mathbf{x}_d$ (i.e., the output of CAP) along with the previous output $\mathbf{y}_{i-1}$ of the side branch are then fed to the following convolution layers and can be written as:
\begin{equation}
    \mathbf{y}_i = \operatorname{Conv}_{\rm PW}(\operatorname{Conv}_{\rm DW}(\operatorname{Conv}_{\rm PW}(\mathbf{x}_{d} + \mathbf{y}_{i-1}))),
    \label{func:lsb_computation}
\end{equation}
where $\operatorname{Conv}_{\rm PW}$ and $\operatorname{Conv}_{\rm DW}$ are point-wise and depth-wise convolution layers, and $i$ is the layer index. In equation~\eqref{func:lsb_computation}, we ignore the normalization and activation layers for simplicity. The structure of LSB is widely used in lightweight models in MobileNet~\cite{howard2017mobilenets,howard2017mobilenets,howard2019searching}, SqueezeNet~\cite{iandola2016squeezenet}, and ShuffleNet~\cite{ma2018shufflenet,zhang2018shufflenet}. We use LSB as a kind of parameter-efficient adapter in the fine-tuning stage. We update all the parameters (i.e., weight and bias) of LSB which requires all the activations. Fortunately, the size of such activation is greatly reduced compared to the main branch. We concatenate the outputs of the backbone and side branch as the final network output.

\subsection{Non-Parametric Layers}
\label{sec:non_linear_layers}
In addition to parametric modules, the non-parametric layers also require the middle features to backpropagate the gradient. Worse yet, this computation cannot be avoided in the backward pass. We seek a solution from quantization to deal with this problem. Neural Network Quantization (NNQ)~\cite{han2015deep,jacob2018quantization,zhao2021distribution,wang2019haq} has long been studied for reducing the storage of parameters and computation complexity in the inference stage. NNQ works on forward propagation and substitutes the full-precision (32-bit) weight and activation with lower bit-width ones. Inspired by this kind of design, we propose to quantize the activation of non-parametric layers for memory efficiency while not changing the computational precision of the non-linear function. 

\begin{figure}
\centering
\includegraphics[width=0.8\textwidth]{    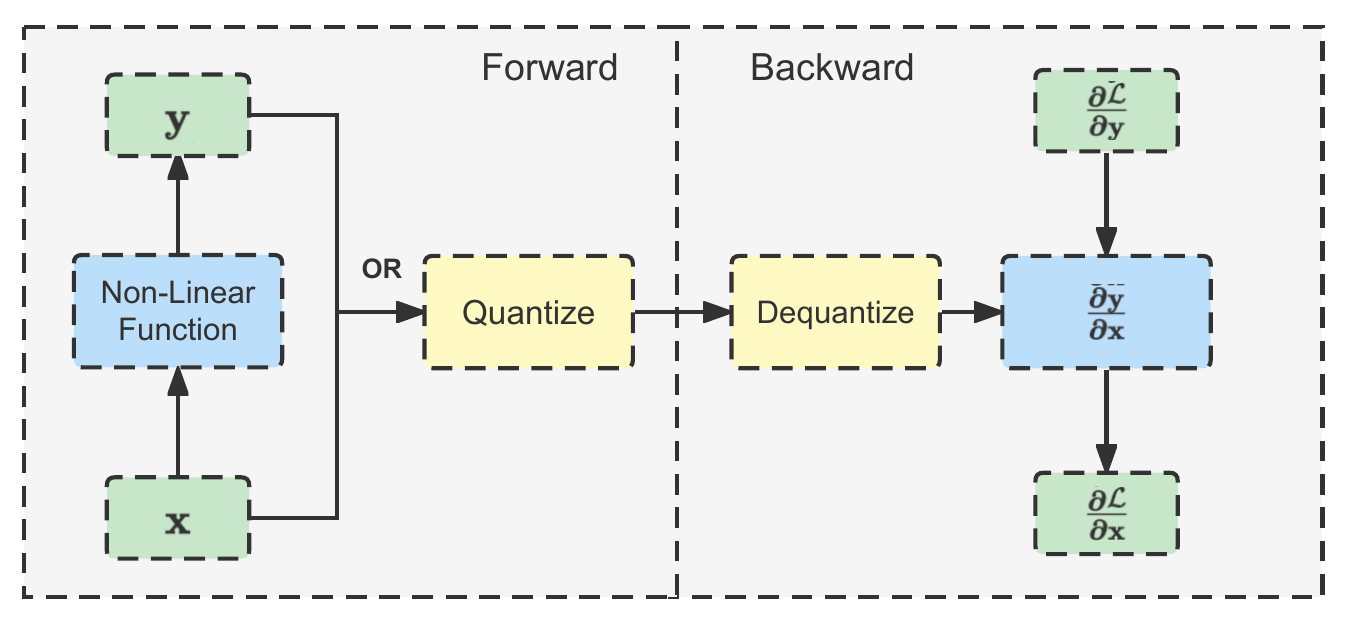}
\caption{Illustration of quantization for non-parametric layers. The activation is quantized to lower bit-width in forward pass and dequantized before propagating the gradient in backward pass.}
\label{fig:nonlinear_quant}
\end{figure}

Figure~\ref{fig:nonlinear_quant} summarizes the quantization process. In the forward pass, we first compute the output $\mathbf{y}$ based on the non-linear function and full-precision input $\mathbf{x}$, then the activation (i.e., $\mathbf{x}$ or $\mathbf{y}$) is quantized to lower bit-width based on the derivative. Thus the quantized activation consumes less memory compared to the full-precision one. In the backward pass, the saved activation is dequantized to the original numerical scale and is then used to compute the gradient of input $\frac{\partial \mathcal{L}}{\partial \mathbf{x}}$ based on $\frac{\partial \mathbf{y}}{\partial \mathbf{x}}$, and the gradient of output $\frac{\partial \mathcal{L}}{\partial \mathbf{y}}$. The detailed implementation varies according to each non-linear function. We discuss the quantization of three widely used non-linear layers (i.e., Softmax, ReLU, and GELU), which are mostly adopted by both CNN (ResNet~\cite{he2016deep}, ConvNext~\cite{liu2022convnet}) and ViT models (ViT~\cite{dosovitskiy2020image}, Swin~\cite{liu2021swin}).

\paragraph{Softmax layer}
Softmax is a vital component in ViTs. For an input vector $\mathbf{x} \in \mathbb{R}^n$ and output vector $\mathbf{y}\in \mathbb{R}^n$, the forward and backward pass can be written as:
\begin{equation}
    \mathbf{y}_i = \frac{e^{\mathbf{x}_i}}{\sum_j e^{\mathbf{x}_j}},\quad \frac{\partial \mathbf{y}_i}{\partial \mathbf{x}_j}= \begin{cases}\mathbf{y}_i\left(1-\mathbf{y}_j\right) & j=i, \\ -\mathbf{y}_i \cdot \mathbf{y}_j & j \neq i.\end{cases}
    \label{func:softmax_computation}
\end{equation}
From equation~\eqref{func:softmax_computation}, we can derive important properties of Softmax. Its gradient can be calculated by $\mathbf{y}_i$ directly and $\sum_i \mathbf{y}_i = 1$. This means the numerical range of $y_i$ is limited to $(0,1)$ and may be significantly below 1, which is ideal for quantization. For example, when the maximum value of $y_i$ is reduced from 1 to 0.1, the numerical precision will be improved by 10 times under the same bits. We use asymmetric quantization~\cite{jacob2018quantization} to quantize the $\mathbf{y}$ to $[0,2^{N}-1]$ during forward where $N$ is the bitwidth, which can be written as:
\begin{equation}
   \mathbf{y}_q = \left\lfloor \frac{\mathbf{y} - \operatorname{min}(\mathbf{y})}{s}\right\rceil, \quad  s = \frac{\operatorname{max}(\mathbf{y}) -\operatorname{min}(\mathbf{y})}{2^{N}-1},
   \label{func:softmax_quantization}
\end{equation}
where $\left\lfloor \cdot \right\rceil$ is a rounding function and $s$ is a scaling factor. During backward, $\mathbf{y}_q$ will be dequantized to the original scale before calculating the gradient,
\begin{equation}
    \hat{\mathbf{y}} = \mathbf{y}_q s + \operatorname{min}(\mathbf{y}),
    \label{func:softmax_dequantization}
\end{equation}
where $\hat{\mathbf{y}}$ is used to calculate the gradient with equation~\eqref{func:softmax_computation}.

\paragraph{ReLU layer}
The derivative of ReLU has only 2 values, i.e., 0 and 1. So the activation $\mathbf{x}$ can be quantized to a binary mask where the location of $\mathbf{x}>0$ requires 1-bit to store.

\begin{figure}[t]
\centering
\includegraphics[width=0.8\textwidth]{    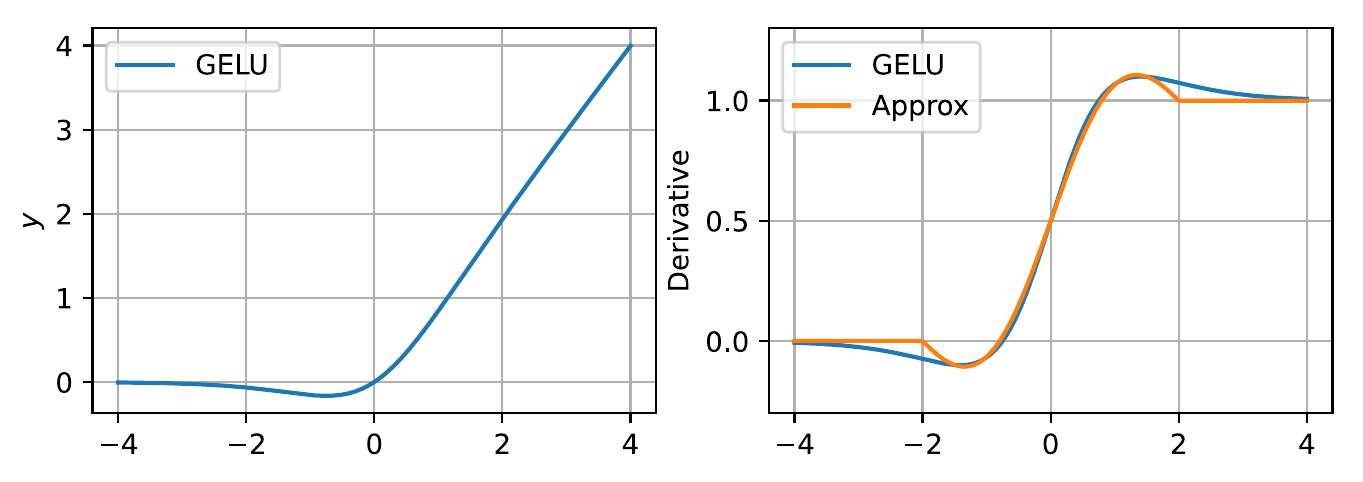}
\caption{Left: the Gaussian Error Linear Unit (GELU); Right: the derivative of GELU and its approximate.}
\label{fig:gelu}
\end{figure}

\paragraph{GeLU layer}
GELU~\cite{hendrycks2016gaussian} is widely used and has shown better performance than ReLU~\cite{liu2022convnet,hendrycks2016gaussian}. The forward and backward propagation of GELU\footnote{we use the approximate proposed by the original paper ~\cite{hendrycks2016gaussian}} can be formulated as,
\begin{equation}
    y = x \operatorname{\sigma}(1.702x),
    \label{func:gelu_forward}
\end{equation}
\begin{equation}
    \frac{\partial y}{\partial x} = \operatorname{\sigma}(1.702x) +1.702xe^{-1.702x} \operatorname{\sigma}^2(1.702x),
    \label{func:gelu_backward}
\end{equation}
where $\operatorname{\sigma}(\cdot)$ is a sigmoid function. There are 2 properties of equation~\eqref{func:gelu_backward}: 1) The second term is an odd function and equal to 0 when $x$ approaches either $-\infty$ or $+\infty$. Second, the function value of equation~\eqref{func:gelu_backward} roughly limited to $(-0.1,1.1)$ and changes faster near the origin while keeping fixed when moving away. This means the quantization error will be relatively smaller when $x$ is away from the origin, and the total quantization error will decrease when reducing the quantization interval. In our case, we quantize the activation of GELU to 4-bit which is sensitive to the quantization range.  Based on these properties, we use one-period sine function to fit the second term of equation~\eqref{func:gelu_backward}. The alternative is shown in Figure \ref{fig:gelu} right side and written as:
\begin{equation}
    \frac{\partial y}{\partial x} = \operatorname{\sigma}(1.702x) + 0.22\operatorname{sin}(1.5x),
    \label{func:gelu_backward_approx}
\end{equation}
where $x$ is first clipped into [-2,2] for smaller quantization error and then quantized between the maximal and minimal value as the same as equation~\eqref{func:softmax_quantization} in the forward pass, except the variable to quantize is the input $\mathbf{x}$ rather than the output $\mathbf{y}$. The quantized variable $\mathbf{x}_q$ can be represented by $N$ bits with set $\{0,1,...,2^N-1\}$ which saves much memory. In the backward pass, $\mathbf{x}_q$ is dequantized by equation~\eqref{func:softmax_dequantization} and fed to equation~\eqref{func:gelu_backward_approx} to calculate the input gradient. The alternative also has less computation complexity compared to that in equation~\eqref{func:gelu_backward}. An additional finding is that our design is also effective for similar non-parametric functions, such as SiLU and Swish~\cite{ramachandran2017searching}.

In sum, we take two strategies towards parameter and memory efficient transfer learning: First, we fine-tune the pretrained model with lightweight activation-efficient modules, i.e., Bias, LRP and LSB. Second, we quantize the required activation in non-parametric layers to lower bit-width based on its derivatives. Different from the methods of NNQ, which quantize the weights for less storage or acceleration in the inference stage, we quantize the activation of non-parametric layers for memory efficiency in fine-tuning stage. 
With these two designs, we can greatly reduce the learnable parameters and training memory in the fine-tuning stage. For example, our S2A reduces the training memory from 4GB to 640MB and only has 0.9\% tunable parameters while keeping competitive performance compared to fully fine-tuning (refer to Section~\ref{sec:main_results} for more details). The quantization for activation is also independent of lite tunable modules proposed in Section \ref{sec:efficient_modules} and can be combined with other tuning methods to reduce the memory footprint. We don't apply quantization to parametric modules in fine-tuning because the numerical range of both required activation and derivative is not limited.

\subsection{Discussion with Prior Arts}
\label{sec:prior_arts}
Our S2A framework shares similarities with prior arts but is significantly different in the memory-efficient design. We discuss the differences upon these works including Adapter~\cite{houlsby2019parameter,chen2022adaptformer,chen2022conv}, LoRA~\cite{hu2021lora}, and VPT~\cite{jia2022visual}.

\paragraph{Adapter} The adapter is proposed by Houlsby et al.~\cite{houlsby2019parameter} for NLP tasks and is extended by Chen et al.~\cite{chen2022adaptformer} for Vision Transformer(ViT).
Each adapter consists of a downsample layer with $\mathbf{W}_{\text {down }}$,
a non-linear activation layer $f(\cdot)$ and a up-sample layer with $\mathbf{W}_{\text {up }}$ such that,
\begin{equation}
    \mathbf{y} = \mathbf{x} +\mathbf{W}_{\text {up }} \cdot f\left( \mathbf{W}_{\text {down }} \cdot \mathbf{x}\right) 
\end{equation}
According to equation~\eqref{func:linear_backward}, the gradient of $\mathbf{W}_{\text {down }}$ relies on the $\mathbf{x}$ which is a full-size activation in backbone, resulting memory non-efficiency. The memory footprint is main caused by updating $\mathbf{W}_{\text {down }}$. Based on this observation, we propose CAP as the downsample layer in our lite side branch. CAP averages each group of the activation along the channel dimension in a parameter-free way. The gradient of input only derives from the gradient of output and the compression factor $r$. Therefore, the storage of full-size activation can be avoided. Under the same compression factor, LSB also has fewer parameters because it keeps the input and output always in the compression space.

\paragraph{VPT}
Prompt tuning has been widely explored in NLP ~\cite{liu2021pre,li2021prefix,lester2021power,liu2021p} and is extended to vision tasks by Jia et al.~\cite{jia2022visual}, Those methods append extra prompt tokens in hidden states. The prompted input of $i$-th layer $L_i$ in the vision transformer can be comprised of class tokens $\mathbf{C}_i$, embeddings $\mathbf{E}_i$, and prompt tokens $\mathbf{P}_i$, which derives,
\begin{equation}
    \mathbf{y} = L_i(\mathbf{x}) = L_i ([\mathbf{C}_i, \mathbf{E}_i,\mathbf{P}_i]).
\end{equation}
The gradient of $\mathbf{P}_i$ is a part of $\frac{\partial \mathcal{L}}{\partial \mathbf{x}}$ and is memory-efficient when the $L_i$ is a linear layer. However, It still has two drawbacks : 1) VPT is difficult to extend to hierarchical or local computing networks, like Swin (the changed feature shape may not be compatible with the pre-trained weights) or CNN (the tokens padded to the feature edge have only local influence). 2) Appending extra tokens to the embedding may seriously aggravate the computational burden when the number of prompt tokens increases, since the computational complexity of the self-attention is $\mathcal{O}(n^2)$ w.r.t the number of tokens in ViT. Our LRP is inspired by the idea of prompting the intermediate activation. We use addition to integrate the low-rank prompt tokens and don't change the feature shape (which means it is easy to extend). Furthermore, LRP doesn't burden the computational cost of the following layers. For the performance comparison between our LRP and VPT, please refer to Section~\ref{sec:expr_analysis}.

\vspace{-5pt}
\paragraph{LoRA}
LoRA decomposes the linear layer into a frozen counterpart $\mathbf{W}$ and two learnable low-rank matrices, i.e., $\mathbf{A}$, $\mathbf{B}$. It only updates $\mathbf{A}$ and $\mathbf{B}$ in fine-tuning and merges the updated parameters to the backbone after training
such that,
\begin{equation}
    \mathbf{y} = \mathbf{x} \mathbf{W} + s\cdot \mathbf{x} \mathbf{AB} = \mathbf{x}(\mathbf{W} + s\cdot \mathbf{A B}),
\end{equation}
where $s$ is a fixed scaling factor. Similar to adapter, the gradient of $\mathbf{A}$ relies on the full-size $\mathbf{x}$ according to equation~\eqref{func:linear_backward}, resulting memory non-efficiency. In contrast, our LRP decomposes the activation (not weight) with two learnable matrices and don't require $\mathbf{x}$ to calculate the gradient.

\vspace{-5pt}
\section{Experiments}

\subsection{Experimental Settings}
\label{sec:experiment_setting}

\noindent\textbf{Pretrained Backbones.}
We introduce two ViTs (i.e., ViT~\cite{dosovitskiy2020image} and Swin~\cite{liu2021swin}) and one CNN backbone (i.e., ConvNeXt~\cite{liu2022convnet}) during experiments. All the backbone are pretrained on ImageNet-21K~\cite{deng2009imagenet} for transfer learning. For ViT, we use the DeiT pretrained backbone~\cite{touvron2021training} without token distillation. 

\noindent\textbf{Benchmark Datasets.}
There are two transfer learning datasets utilized in our experiments. The first one is the VTAB-1K~\cite{zhai2019large} which consists of 19 diverse image classification datasets. These datasets are divided into 3 group1s (i.e., Natural, Specialized, and Structured) with 7, 4, 8 subsets respectively. Each dataset contains 1000 training images. The second dataset is FGVC which contains 8 fine-grained image classification datasets including CUB-200-2011\cite{WahCUB_200_2011}, Oxford Flowers\cite{nilsback2008automated}, Stanford Dogs\cite{khosla2011novel}, Caltech101\cite{fei2006one}, DTD\cite{cimpoi2014describing}, EuroSAT\cite{helber2019eurosat}, Oxford Pets\cite{parkhi2012cats}, and UCF101\cite{soomro2012ucf101}. We follow~\cite{jia2022visual} for training configurations. We report the average accuracy of these two datasets in the main paper and show the results of the 27 datasets in the supplementary materials.

\noindent\textbf{Compared Methods.}
We compare our S2A with several state-of-the-art fine-tuning methods, including
1) \texttt{Full}: all the parameters are learnable during tuning. 2)  \texttt{Linear}: only tune the last linear layer as a classification head. 3) \texttt{Adapter}: following the setting in \cite{houlsby2019parameter}, two learnable adapters are inserted into each block of the backbone. In ViT and Swin, the adapter comprises fully-connected layers, while the counterparts are substituted by convolutions in ConvNeXt for compatibility.   4) \texttt{LoRA}\cite{hu2021lora}: add parallel low-rank linear layers in self-attention. 5) \texttt{VPT}\cite{jia2022visual}: add extra tokens in activation where the VPT-deep version is used. Unless specified, we only tune the introduced modules or parameters of the above methods. 

\noindent\textbf{Implementation Details.}
Our S2A is composed of three parametric modules (i.e., LRP, LSB, and Bias). The features of the main branch are downsampled by $8\times$ along the channel dimension and then fed to LSB. The scale factor $r$ in LRP is fixed at 30. For comparison, the number of prompt tokens in VPT is set to 50, and the features in LoRA and Adapter are downsampled by 8$\times$. For non-parametric layers (i.e., Softmax and GELU in ViT and Swin backbone, GELU in ConvNeXt backbone), we save the 4-bit quantized activation to calculate the gradients in the backward pass. We report the percentage of tunable parameters and training memory footprint under a batch size of 32 for each method. All the experiments are conducted via Pytorch~\cite{paszke2019pytorch} on a single Nvidia V100 GPU. 
Additionally, we use the theoretically calculated training memory footprint for comparison in our experiments since PyTorch does not support explicit fine-grained memory management.
We use AdamW optimizer\cite{loshchilov2018fixing} and cosine decay learning rate schedule\cite{loshchilov2016sgdr} with 10 epochs for warming up and 70 epochs for training.

\begin{table}[htbp]
\small
\centering
\setlength\tabcolsep{4pt}
\resizebox{0.8\textwidth}{!}{%
\begin{tabular}{@{}cccccccc@{}}
\toprule
\multirow{2}{*}{} & \multirow{2}{*}{\begin{tabular}[c]{@{}c@{}}Params\\ (\%)\end{tabular}} & \multirow{2}{*}{Quantization} & \multirow{2}{*}{\begin{tabular}[c]{@{}c@{}}Memory\\ (MB)\end{tabular}} & \multicolumn{3}{c}{VTAB-1k}        & \multirow{2}{*}{FGVC} \\ \cmidrule(lr){5-7}
                  &                                                                       &                         &                                                                      & \scriptsize Natural & \scriptsize Specialized & \scriptsize Structured &                       \\ \toprule
\multicolumn{8}{c}{ViT-B/16}                                                                                                                                                                                                                            \\ \midrule
Full              & 100\%                                                                 &                         & 4099                                                                 & 74.94   & 85.68       & 51.12      & 88.52                 \\
Linear            & 0.09\%                                                                &                         & 344                                                                  & 68.26   & 82.16       & 35.50      & 85.77                 \\
LoRA              & 2.15\%                                                                &                         & 2242                                                                 & 76.06   & 84.82       & 47.99      & \textbf{89.38}                 \\
Adapter           & 4.24\%                                                                &                         & 2583                                                                 & 74.99   & 84.81       & 50.06      & 87.37                 \\
VPT               & 0.63\%                                                                &                         & 2453                                                                 & 76.33   & 84.75       & 49.82      & 89.12                 \\
UniPT               & 2.96\%                                                                &                         & 1472                                                                 & 76.45   & \textbf{85.73}       & 50.63      & 89.24                 \\
\rowcolor{LightBlue} S2A               & 0.90\%                                                                & $\checkmark$            & 640                                                                  & \textbf{76.62}   & 84.38       & \textbf{51.94}     & 88.87                 \\ \toprule
\multicolumn{8}{c}{Swin-B}                                                                                                                                                                                                                              \\ \midrule
Full              & 100\%                                                                 &                         & 6390                                                                 & 80.02   & 86.81       & \textbf{57.39}      & 90.83                 \\
Linear            & 0.12\%                                                                &                         & 353                                                                  & 74.57   & 82.49       & 33.75      & 90.15                 \\
Adapter           & 2.15\%                                                                &                         & 3228                                                                 & 80.86   & \textbf{86.89}       & 53.38      & 91.17                 \\
VPT               & 0.79\%                                                                &                         & 3952                                                                 & 80.94   & 84.43       & 45.23      & 91.34                 \\
UniPT               & 1.47\%                                                                &                         & 2068                                                                 & 81.11   & 84.94       & 52.74      & 90.49                 \\
\rowcolor{LightBlue} S2A               & 1.03\%                                                                & $\checkmark$            & 745                                                                  & \textbf{81.89}   & 85.73       & 56.87      & \textbf{92.12}                 \\ \toprule
\multicolumn{8}{c}{ConvNeXt-B}                                                                                                                                                                                                                          \\ \midrule
Full              & 100\%                                                                 &                         & 7171                                                                 & 81.10   & \textbf{86.71}       & 51.26      & 91.36                 \\
Linear            & 0.11\%                                                                &                         & 350                                                                  & 75.34   & 83.58       & 35.48      & 91.09                 \\
Adapter           & 2.0\%                                                                 &                         & 3743                                                                 & 80.08   & 85.53       & 41.01      & 90.97                 \\
VPT               & 1.28\%                                                                &                         & 3265                                                                 & 77.66   & 83.83       & 39.32      & 91.88                 \\
UniPT               & 1.63\%                                                                &                         & 2783                                                                 & 82.04   & 86.29       & 54.68      & 91.96                 \\
\rowcolor{LightBlue}S2A               & 0.98\%                                                                & $\checkmark$            & 741                                                                  & \textbf{82.63}   & 85.60       & \textbf{55.87}      & \textbf{92.04}                 \\ \bottomrule
\end{tabular}
}
\caption{Performance comparison on different downstream tasks with three pre-trained backbones, i.e., ViT-B/16, Swin-B, ConvNeXt-B. Our methods show appealing results compared to SOTA methods while significantly reducing the training memory.}
\label{tab:main_results}
\end{table}

\begin{table}[htbp]
\centering
\setlength\tabcolsep{4pt}
\resizebox{0.8\textwidth}{!}{%
\begin{tabular}{cccccc}
\toprule
\multirow{2}{*}{Quantization} & \multirow{2}{*}{\begin{tabular}[c]{@{}c@{}}Memory\\ (MB)\end{tabular}} & \multicolumn{3}{c}{VTAB-1k}        & \multirow{2}{*}{FGVC} \\ \cline{3-5}
                  &                                                                       & \scriptsize Natural & \scriptsize Specialized & \scriptsize Structured &                       \\ \toprule
\multicolumn{6}{c}{ViT-B/16}                                                                                                                           \\ \hline
Unquantized               & 2097                                                                  & 76.55   & \textbf{84.54}       & \textbf{52.29}      & \textbf{89.10}                 \\
\rowcolor{LightBlue}\small Softmax\&GELU               & 640                                                                  & \textbf{76.62}   & 84.38       & 51.94      & 88.87                 \\ \toprule
\multicolumn{6}{c}{Swin-B}                                                                                                                             \\ \hline
Unquantized               & 2672                                                                  & \textbf{82.23}   & 85.38       & 56.57      & \textbf{92.13}                 \\
\rowcolor{LightBlue}\small Softmax\&GELU               & 745                                                                  &81.89   & \textbf{85.73}       & \textbf{56.87}      & 92.12                 \\ \toprule
\multicolumn{6}{c}{ConvNeXt-B}                                                                                                                         \\ \hline
Unquantized               & 2832                                                                  & \textbf{82.68}   & \textbf{85.70}       & 55.52      & \textbf{92.08}                 \\
\rowcolor{LightBlue}GELU               & 741                                                                  & 82.63   & 85.60       & \textbf{55.87}      & 92.04                 \\ \bottomrule
\end{tabular}
}
\caption{Effect of the quantization on GELU and Softmax. For each pretrained backbone, the tasks of first row are full-precision non-parametric layers which are quantized to 4-bit in the second row. The results show relative small change on performance.}
\label{tab:quant_comparison}
\end{table}

\subsection{State-of-the-art Comparisons}
\label{sec:main_results}

Table~\ref{tab:main_results} shows our comparison to state-of-the-art PETL methods. Under the ViT-B/16 backbone, our S2A achieves the best accuracy under 2 groups of VTAB-1K dataset. Under the Swin-B backbone, our S2A performs best in 1 group of VTAB-1K and the FGVC dataset. Under the ConvNext-B backbone, our S2A performs best in most datasets. Especially, the VTAB-1K `structured' group has less domain affinity to the pre-trained ImageNet dataset~\cite{evci2022head2toe}. Therefore it is more difficult to adapt, while our S2A shows relatively large improvement in this group.
Overall, Our S2A shows competitive performance compared to other PETL methods including the full finetuning results. 

Besides classification accuracy, our S2A is both parameter and memory efficient. Under these three backbone, the training memory of S2A is 4$\times$ smaller than VPT, Adapter, and Lora on average. Compared to the full-finetuing, S2A reduces training memory by 6.4$\times$, 8.6$\times$ and 9.7$\times$ on ViT-B/16, Swin-B, and ConvNeXt-B, respectively. The significant memory reduction is because of our memory-efficient module design and quantized activations for non-parametric layers.

\subsection{Experimental Analysis.}
\label{sec:expr_analysis}

\paragraph{Quantized Non-Parametric Layers}
We evaluate our quantization by comparing it to the unquantized design, which is the full-precision storage adopted by existing PETL methods. The activations of non-parametric layers with their memory-efficient counterparts are quantized to 4-bit. Table~\ref{tab:quant_comparison} shows the results. In most datasets, the quantized layers (on the second rows) have little accuracy degradation (i.e., below 0.4\%) compared to the unquantized ones. Moreover, in some datasets the quantized results improve upon unquantized ones (e.g., Swin-B on `Specialized' and `Structured', ConvNeXt-B on `Structured'). On the memory side, the quantization reduces 3$\times$ memory overall, and the memory of non-parameterized layers by 8$\times$ as we use 4-bit to substitute 32-bit full precisions. In sum, the little accuracy change between quantization and dequantization is due to the properties of Softmax and GELU. For Softmax, the value range of the required activation is limited to a small interval. For GELU, the derivative is roughly limited to $(-0.1, 1.1)$ and changes trim when moving away from the origin.

\vspace{-5pt}
\paragraph{LRP v.s. VPT}
Our LRP design shares similarities to VPT at a glimpse while containing significant differences. Our LRP directly adds the prompt tokens to the embeddings after two low-rank matrix multiplications. In contrast, VPT concates extra prompt tokens to the hidden state embeddings, which will change the input shape and may not be compatible with the pretrained weights. Such concatenation design is difficult for hierarchical ViT (e.g., Swin) or CNNs, while our addition design does not suffer. Table~\ref{tab:comparison_lrp_vpt} shows the comparing results. The accuracy of LRP is higher on Swin-B and ConvNeXt-B, while it is lower on ViT-B. The lower accuracy of LRP on ViT is that VPT concatenates tokens to the original embeddings to increase self-attention potentials, while LRP only performs addition to preserve the embedding sizes. On the other hand, LRP performs better on Swin-B and ConvNeXt-B. This is due to VPT that only applies to the local window attention because of the shape compatibility in Swin, and pads the edge of features resulting in regional influence in CNNs. In future work, we will further explore how to suit the prompt token embedding in a more extendable and effective way.

\begin{table}[htbp]
\centering
\resizebox{0.6\textwidth}{!}{%

\begin{tabular}{cccccc}
\toprule
\multirow{2}{*}{} & \multirow{2}{*}{\begin{tabular}[c]{@{}c@{}}Parameters\\ (M)\end{tabular}} & \multicolumn{3}{c}{VTAB-1k}        & \multirow{2}{*}{FGVC} \\ \cline{3-5}
                  &                                                                       & \scriptsize Natural & \scriptsize Specialized & \scriptsize Structured &                       \\ \toprule
\multicolumn{6}{c}{ViT-B/16}                                                                                                                           \\ \hline
VPT               & 0.54                                                                  & \textbf{76.33}   & \textbf{84.75}       & \textbf{49.82}      & \textbf{89.12}                 \\
\rowcolor{LightBlue}LRP               & 0.54                                                                  & 75.48   & 83.18       & 48.86      & 88.64                 \\ \toprule
\multicolumn{6}{c}{Swin-B}                                                                                                                             \\ \hline
VPT               & 0.69                                                                  & 80.94   & 84.43       & 45.23      & 91.34                 \\
\rowcolor{LightBlue}LRP               & 0.61                                                                  & \textbf{81.32}   & \textbf{84.97}       & \textbf{49.13}      & \textbf{91.62}                 \\ \toprule
\multicolumn{6}{c}{ConvNeXt-B}                                                                                                                         \\ \hline
VPT               & 1.12                                                                  & 77.66   & 83.83       & 39.32      & 91.88                 \\
\rowcolor{LightBlue}LRP               & 0.74                                                                  & \textbf{81.29}   & \textbf{85.12}       & \textbf{50.05}      & \textbf{92.00}                 \\ \bottomrule
\end{tabular}
}
\caption{Comparison between LRP and VPT. LRP has better performance on Swin-B and ConvNeXt-B.}
\label{tab:comparison_lrp_vpt}
\end{table}

\paragraph{Ablation Studies}
Our S2A consists of three parameter and memory-efficient modules (i.e., Bias, LRP, and LSB). We analyze how these modules work together. Initially, we only apply tunable bias in the backbone. Then, we add LRP and LSB to show the performance change. Table~\ref{tab:components_S2A} shows the results where adding LRP and LSB improves the original bias tuning performance. Moreover, the performance is improved more obviously on the `Structured' group of VTAB-1K (i.e., 6.63\% under ViT-B, 8.26\% under Swin-B, and 5.06\% under ConvNeXt-B). Ignoring a slight degradation on 3 tasks, tuning the Bias, LRP and LSB together shows the most capacity for transfer learning.

\begin{table}[h]
\centering
\resizebox{0.6\textwidth}{!}{%

\begin{tabular}{ccccc}
\toprule
\multirow{2}{*}{} & \multicolumn{3}{c}{VTAB-1k}        & \multirow{2}{*}{FGVC} \\ \cline{2-4}
                  & \scriptsize Natural & \scriptsize Specialized & \scriptsize Structured &                       \\ \toprule
\multicolumn{5}{c}{ViT-b/16}                                                        \\ \hline
Bias              & 75.38   & 83.41       & 45.66      & 87.58                 \\
Bias+LRP          & 76.29   & 83.72       & 49.53      & 88.61                 \\
\rowcolor{LightBlue}Bias+LRP+LSB      & \textbf{76.55}   & \textbf{84.54}       & \textbf{52.29}      & \textbf{89.10}                 \\ \toprule
\multicolumn{5}{c}{Swin-B}                                                     \\ \hline
Bias              & 80.73   & 85.18       & 48.31      & 90.59                 \\
Bias+LRP          & 82.12   & \textbf{85.99}       & 53.67      & 91.82                 \\
\rowcolor{LightBlue}Bias+LRP+LSB      & \textbf{82.23}   & 85.38       & \textbf{56.57}      & \textbf{92.13}                 \\ \toprule
\multicolumn{5}{c}{ConvNeXt-B}                                                 \\ \hline
Bias              & 80.19   & 83.86       & 50.46      & 91.15                 \\
Bias+LRP          & \textbf{82.74}   & 85.53       & 53.78      & \textbf{92.10}                 \\
\rowcolor{LightBlue}Bias+LRP+LSB      & 82.68   & \textbf{85.70}       & \textbf{55.52}      & 92.08                 \\ \bottomrule
\end{tabular}
}
\caption{Comparison between different components. We gradually add those three components to the pretrained backbone and then fine-tune in VTAB-1k and FGVC datasets. The best results are obtained when those three components are combined.}
\vspace{-12pt}
\label{tab:components_S2A}
\end{table}

\begin{figure}[htbp]
\centering
\includegraphics[width=0.8\textwidth]{    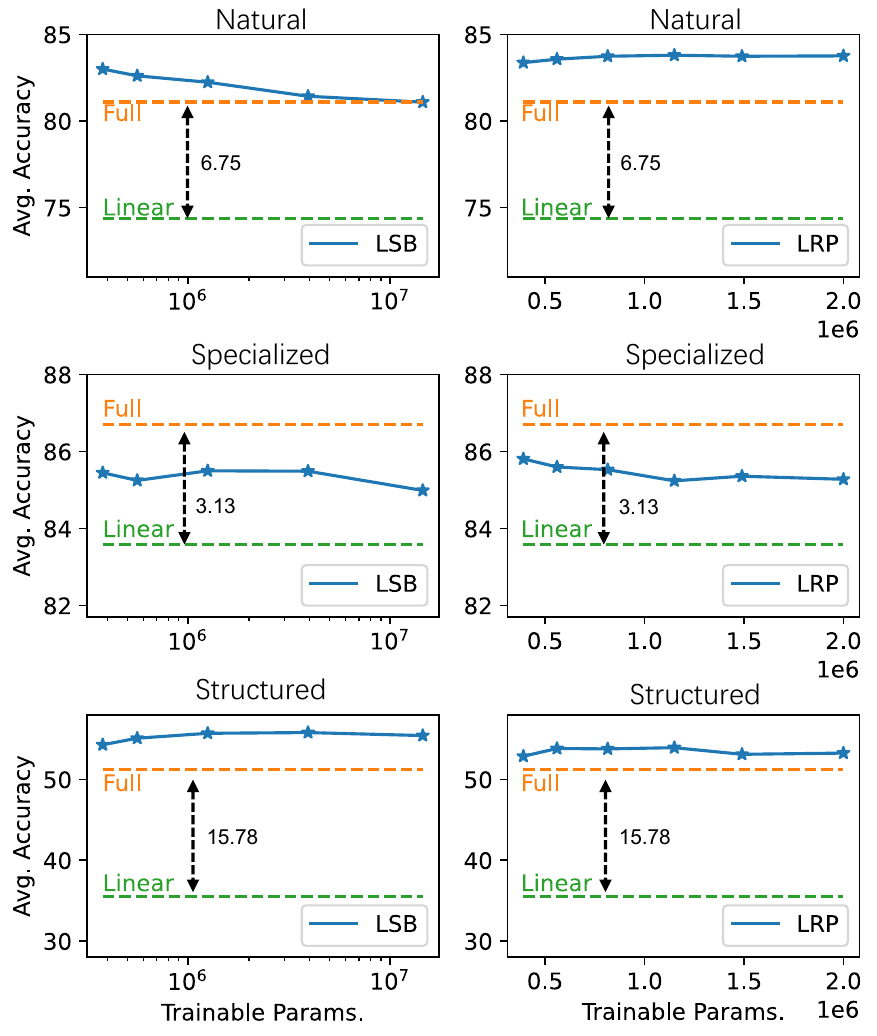}
\caption{The trend of performance on ConvNeXt-B as the amount of learnable parameters grows up.}
\label{fig:scale_up_params}
\end{figure}

\vspace{-3pt}

\paragraph{Scaling Tunable Parameters Up}
We analyze how the number of learnable parameters of S2A affects on the transfer performance. We use ConvNeXt-B as the backbone and divide S2A into two parts (i.e., LRP and LSB), and gradually increase the tunable parameters to observe the performance changes. For LRP, we set the scaling factor $r$ as 5, 15, 30, 50, 70, and 100, respectively. For LSB, we downsample the features by 16$\times$, 8$\times$, 4$\times$, 2$\times$, and 1$\times$, respectively. Figure~\ref{fig:scale_up_params} shows the evaluation results. The performance of LSB gradually decreases in the `Natural' group of VTAB-1K and increases a little in the `Structured' group. This is due to the domain affinity~\cite{evci2022head2toe} which can be measured by the performance difference between \texttt{Full} and \texttt{Linear} fine-tuning. In the `Natural' group, the data resembles the pretraining dataset (ImageNet) more than that in the `Structured' group. So an increasing performance can be obtained by tuning fewer parameters in the `Natural' group. As the randomly initialized parameters scale up in LSB, the model may be under-fitting on small datasets, resulting in performance degradation. This effect is not obvious on `Structured' because the difference between the pre-training and fine-tuning domains is relatively large. LRP has more stability on 3 groups of VTAB-1k compared to LSB, because the model is smoothly transferred from source domain to target domain by initializing the added tokens to 0.

\vspace{-5pt}
\section{Conclusion}
\vspace{-3pt}
In this work, we propose S2A towards a unified view of parameter and memory-efficient transfer learning. Unlike the previous PETL methods only take the storage of fine-tuned parameters into account, we further try to reduce the training memory in fine-tuning stage from two perspectives, 1) we derive the memory-efficient parametric modules based on the requirement for the activation. 2) we quantize the activation of non-parametric layers into lower bit-width based on the properties of their derivative. Our S2A shows competitive performance compared to fully fine-tuning, e.g., 71.99\%(S2A) vs. 69.72\%(Full) on the VTAB-1k dataset and ConvNeXt-B. At the same time, the amount of tunable parameters is reduced by over 100$\times$ and the training memory is reduced by 9.7$\times$. The experimental performance shows the promising prospects of our S2A framework on both training and practical deployment.
\clearpage

\bibliographystyle{unsrt}

    \bibliography{template}

\end{document}